\documentclass[]{article}
\usepackage[letterpaper]{geometry}
\usepackage{mtsummit2021}
\usepackage{times}
\usepackage{url}
\usepackage{latexsym}
\usepackage{natbib}
\usepackage{layout}

\usepackage{color}
\usepackage{amsmath}
\usepackage{bm}
\usepackage{booktabs}
\usepackage{graphicx}
\usepackage{CJKutf8}
\usepackage{tabularx}
\newcommand{\ja}[1]{\begin{CJK}{UTF8}{min}#1\end{CJK}}
\newcommand{\minisection}[1]{\noindent{\bf {#1}.}}
\usepackage{cleveref}
\crefformat{section}{\S#2#1#3}
\crefformat{subsection}{\S#2#1#3}
\crefformat{subsubsection}{\S#2#1#3}
\usepackage{multirow}

\newcommand{\fig}[1]{Fig. \ref{#1}}
\newcommand{\eq}[1]{Eq. \ref{#1}}

\begin{document}

\title{\bf Modeling Target-side Inflection in \\Placeholder Translation}

\author{\name{\bf Ryokan Ri} \hfill  \addr{li0123@logos.t.u-tokyo.ac.jp}\\
        \name{\bf Toshiaki Nakazawa} \hfill \addr{nakazawa@logos.t.u-tokyo.ac.jp}\\
       \name{\bf Yoshimasa Tsuruoka} \hfill \addr{tsuruoka@logos.t.u-tokyo.ac.jp}\\
        \addr{The University of Tokyo, 7-3-1 Hongo, Bunkyo-ku, Tokyo, Japan}
}


\maketitle
\pagestyle{empty}

\begin{abstract}
Placeholder translation systems enable the users to specify how a specific phrase is translated in the output sentence.
The system is trained to output special placeholder tokens, and the user-specified term is injected into the output through the context-free replacement of the placeholder token.
However, this approach could result in ungrammatical sentences because it is often the case that the specified term needs to be inflected according to the context of the output, which is unknown before the translation.
To address this problem, we propose a novel method of placeholder translation that can inflect specified terms according to the grammatical construction of the output sentence.
We extend the sequence-to-sequence architecture with a character-level decoder that takes the lemma of a user-specified term and the words generated from the word-level decoder to output the correct inflected form of the lemma.
We evaluate our approach with a Japanese-to-English translation task in the scientific writing domain, and show that our model can incorporate specified terms in the correct form more successfully than other comparable models.\footnote{Code is available at \url{https://github.com/Ryou0634/placeholder_translation}.}

\end{abstract}

\section{Introduction}
Over the last several years, neural machine translation (NMT) has pushed the quality of machine translation to near-human performance \citep{NIPS2014_a14ac55a,NIPS2017_3f5ee243}.
However, due to its end-to-end nature, this comes with the cost of losing a certain degree of control over the produced translation, which once was explicitly modeled, for example, in the form of phrase table \citep{koehn-etal-2003-statistical} in statistical machine translation (SMT).
In practice, users often want to specify how certain words are translated in order to ensure the consistency of document-level translation or to guarantee the model to produce the correct translation for words that may be underrepresented in the training corpus such as proper nouns, technical terms, or novel words.

Given this motivation, a line of previous research has investigated {\it placeholder translation} \citep{post-etal-2019-exploration}.
With a source sentence where certain words are replaced with a special placeholder token, the model produces a translation with the special placeholder token in an appropriate position, and then that placeholder token is replaced with a pre-specified term in a post-processing step.

Although this approach ensures that certain words appear in the translation, one limitation is that the user must specify the term that fits in the context surrounding the placeholder token, or specifically, the term should be properly inflected according to the syntactic structure of the produced translation.
To illustrate the problem, we show an actual output from a normal placeholder translation model in Japanese to English translation in Table \ref{fig:motivation_example}.

The system is supposed to translate the word \ja{管理} into {\it controlling} as in the reference, but the output has a different grammatical construction and thus the progressive form {\it controlling} is invalid in this context; instead, {\it controlled} should be injected in the placeholder.
The appropriate word form is difficult to predict, especially in translation between grammatically distant languages, such as Japanese and English. As manually correcting the inflection in post-editing significantly hurts the convenience of placeholder translation, we need a way to automatically handle inflection.

\begin{table}[t]
  \centering
  \begin{tabularx}{\textwidth}{X} \toprule
  {\bf Specified Translation}: \textcolor{red}{\ja{管理} $\rightarrow$ controlling} \\
  {\bf Source}: \ja{フローセンサーの原理は浮遊式流量計のテーパー管内フロートの位置を差動トランスで検出し,これの電圧制御により流量を\textcolor{red}{[VERB]}する。}\\
  {\bf Reference}: The sensor controls the flow rate by detecting the position of the float in the tepered tube with a a differential transformer and \textcolor{red}{[VERB]} it with the obtained voltage. \\
  {\bf System Output}: The principle of the flow sensor is that the position of the float in the taper tube of the floating flowmeter is detected by the differential transformer, and the flow rate is \textcolor{red}{[VERB]} by this voltage control. \\ \bottomrule
  \end{tabularx}
\caption{A translation example from the ASPEC corpus \citep{nakazawa-etal-2016-aspec} with a placeholder translation model. The specified target term grammatically fits the placeholder in the reference, but not in the system output as it is.}
\label{fig:motivation_example}
\end{table}

One possible approach to this problem is the code-switching methods, in which certain words in the source sentence are replaced with the specific target words, and the model is encouraged to include those specific words in the translation. This approach is flexible in that the model can inflect the specified words according to the context \citep{song-etal-2019-code}, but less faithful to the lexical constraints, often ignoring the specified terms (\cref{sec:results}).

To address this problem, we propose a model that automatically inflects a pre-specified term according to the context of the produced translation.
We extend the sequence-to-sequence encoder and decoder with an additional character-level decoder that predicts the inflected form of the pre-specified term.
Our approach combines the advantages of both the placeholder and the code-switching methods: the faithfulness to lexical constraints and the flexibility of dynamically deciding the word form in the output.

We test our approach with a Japanese-to-English translation task in the scientific-writing domain \citep{nakazawa-etal-2016-aspec}, where the translation of technical terms poses a challenge to a vanilla NMT system.
The results show that the proposed method can include the specified term in the appropriately inflected form in the translation with higher accuracy than a comparable code-switching method.
We also perform a careful error analysis to understand the weaknesses of each system and suggest directions for future work.

\section{Related Work}
\subsection{Placeholder Translation}
To ensure that certain words appear in the translated sentence, previous studies have explored the method of replacing certain classes of words with special placeholder tokens and restore the words in a post-processing step, which we call {\it placeholder translation} in this paper.

\citet{luong-etal-2015-addressing} and \citet{long-etal-2016-translation} employed placeholder tokens to improve the translation of rare words or technical terms.
However, simply replacing words with a unique placeholder token loses the information on the original words. To alleviate this problem, subsequent studies distinguish different types of placeholders, such as named entity types \citep{Crego2016SYSTRANsPN,post-etal-2019-exploration} or parts-of-speech \citep{michon-etal-2020-integrating}.

Instead of replacing the placeholder token with a dictionary entry, some studies propose generating the content of the placeholder with a character-level sequence-to-sequence model to translate words not covered in the bilingual dictionary. \citet{Li2016NeuralNT} and \citet{wang-etal-2017-sogou} incorporated a named entity translator, which is supposed to learn transliteration of named entities.
As in their work, our proposed model also uses a character-level decoder to generate the content of placeholders, but our focus is to inflect a lemma to the appropriately inflected form given the context.

\subsection{The Code-switching Method}
Another way to introduce terminology constraints is the code-switching method \citep{song-etal-2019-code,dinu-etal-2019-training,exel-etal-2020-terminology}. The model is trained with source sentences where some words are replaced or followed by specific target words and expected to copy the words to the translation.

One advantage of the code-switching method is that, unlike the placeholder methods, it preserves the meaning of the original words, which likely leads to better translation quality.
Also, the model can incorporate the specified terminology in a flexible way: a model trained with the code-switching method not only copies the pre-specified target words but can inflect the words according to the target-side context \citep{dinu-etal-2019-training}.
In parallel to our work, \citet{niehues-2021-continuous} offers a quantitative evaluation of how well the code-switching method handles inflection of a pre-specified terminology when the terminology is given in the lemma form.

Although the code-switching method is flexible, one disadvantage is that it tends to ignore the pre-specified terminology more often than the placeholder method (\cref{sec:results}).
We propose a placeholder method that handles inflection of pre-specified terms, aiming for both flexibility and faithfulness to terminology constraints.

\subsection{Constrained Decoding}
Another approach to ensure that a pre-specified term appears in the translation is constrained decoding \citep{anderson-etal-2017-guided,hokamp-liu-2017-lexically,post-vilar-2018-fast}.
Constrained decoding can be applied to any existing NMT models without modifying its architecture and training regime, but imposes a significant cost on the decoding speed.
It is also unclear how to incorporate lexical inflection into constrained decoding.
Therefore, we focus on the placeholder and code-switching methods in this study.

\subsection{Modeling Morphological Inflection in Neural Machine Translation}
Explicitly modeling morphological inflection into NMT models has been studied mainly to enable effective generalization over morphological variation of words.
\citet{tamchyna-etal-2017-modeling} and \citet{weller-di-marco-fraser-2020-modeling} propose to decompose certain classes of words into its lemma and morphological tags to reduce data sparsity.
At decoding time, the inflected form is restored by a morphological analyzer. \citet{Song_Zhang_Zhang_Luo_2018} proposed a model that only requires a stemmer to alleviate the need for linguistic analyzers. The model decomposes the process of word decoding into stem generation and suffix prediction.

In this work, we propose to model morphological inflection in the process of embedding pre-specified terms into placeholders to improve the flexibility of placeholder translation. Our approach requires no external linguistic analyzer at prediction time; instead, inflection is performed via a neural character-based decoder.

\section{Approach}
\label{sec:approach}
The proposed model builds upon a sequence-to-sequence (seq2seq) model with an attention mechanism.
Specifically, we use the Transformer model \citep{NIPS2017_3f5ee243}.

In the normal placeholder translation, the model is trained to generate placeholder tokens \texttt{[PLACEHOLDER]} when the source sentence includes them.
Then the placeholder tokens are replaced with user-provided terms in post-processing.

We extend the model to be able to handle inflection. Specifically, we consider the scenario where lemmas are provided as a specified term.
On top of the (sub)word-level decoder, we stack a character-level decoder to generate the content of the placeholder token.
The character-level decoder has to predict the correct inflected form of the specified lemma in the surrounding context.
Specifically, given the target tokens $\{w_1,..., w_T\}$ that contain a placeholder token and the specified lemma that consists of $L$ characters $c_{lemma} = \{c_1, .., c_L\}$, the character decoder generates the inflected form $c_{infl} = \{c'_1, .., c'_{L'}\}$.

\begin{figure}[t]
\centering
\includegraphics[width=14.0cm]{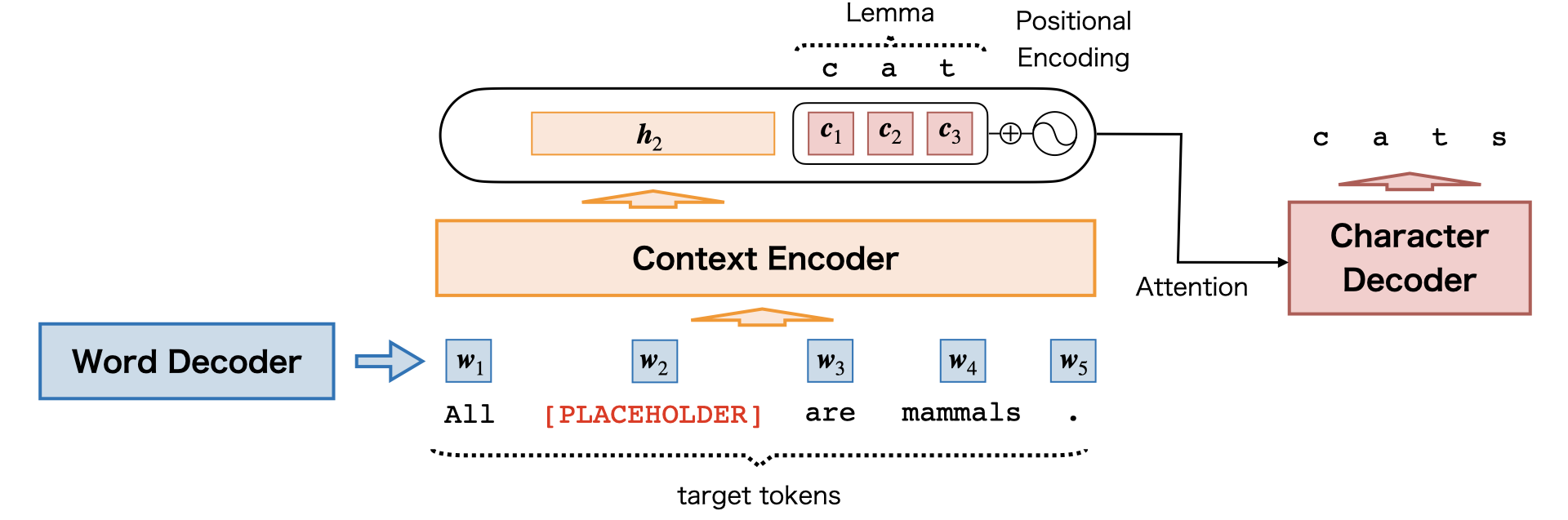}
\caption{The proposed method: placeholder translation with a character decoder.}
\label{fig:architecture}
\end{figure}

We model the generation process with a decoder with attention mechanism (\fig{fig:architecture}).
We first summarize the contextual information on the placeholder token by a context encoder.
Specifically, we feed the embeddings of the target tokens $\{\bm{w}_1,..., \bm{w}_T\}$ into another Transformer encoder to contextualize the placeholder token (\eq{eq1}).
Then, the contextualized  representation of the placeholder token $\bm{h}_p$ and the character embeddings of the specified lemma $\{\bm{c}_1, .., \bm{c}_L\}$ with positional encoding \citep{NIPS2017_3f5ee243} are concatenated to form key-value vectors for decoder attention (\eq{eq2}).
Finally, the key-value vectors are passed to the character-level Transformer decoder and it generates the inflected form $\{c'_1, .., c'_{L'}\}$ in an auto-regressive manner (\eq{eq3}).

\begin{align}
  & \bm{h}_{1},..., \bm{h}_{T} = \operatorname{ContextEncoder}([\bm{w}_1,..., \bm{w}_T]) \label{eq1} \\
  & \bm{A} = [\bm{h}_p; \operatorname{Positional}(\bm{c}_1, .., \bm{c}_L)] \text{  where  } w_p = \texttt{[PLACEHOLDER]} \label{eq2} \\
  & c'_{t} = \operatorname{CharacterDecoder}(\bm{c'}_{<t}, \bm{A}) \label{eq3}
\end{align}

\section{Experimental Setups}
We evaluate the proposed model with several baselines to show how well the model can produce the appropriately inflected form of a given lemma.

\subsection{Corpus}
We conduct experiments in a Japanese-to-English translation task with the ASPEC corpus \citep{nakazawa-etal-2016-aspec}.
This corpus consists of abstracts from scientific articles, which tend to contain many technical terms.
Such words are rare and hard for the model to learn the correct translation, and thus this corpus fits the typical use-case of lexically constrained translation.
We use the initial 1M sentence pairs from the training split for training.

\subsection{Word Dictionary}
In this study, lexical constraints in translation are introduced through a source-to-target word dictionary. We construct the dictionary automatically from the ASPEC corpus through the following procedure.

First, we obtain the word alignment by feeding the first 1M sentence pairs of the training split and validation/test splits to \texttt{GIZA++}.\footnote{\url{https://github.com/moses-smt/giza-pp}} We tokenize Japanese sentences with \texttt{Mecab}\footnote{\url{https://taku910.github.io/mecab/}} and English sentences with \texttt{spaCy}.\footnote{\url{https://spacy.io/}} We then construct a phrase table and extract only those with more than 100 occurrences.
Then, we split the dictionary into noun and verb entries to facilitate the analysis of the results and remove noise. If both the Japanese and English phrases are noun phrases, the entry is registered in the noun dictionary. If the Japanese phrase is a nominal verb\footnote{The nominal verb (\ja{サ変動詞}) is the most productive class of verb in Japanese and many new or technical terms fall into this category ({\it e.g.}, \ja{最適化する}-{\it optimize}, \ja{過学習する}-{\it overfit}).} and English is a verb, the entry is registered in the verb dictionary.
In this study, we evaluate the model's ability to inflect a provided lemma. Lemmas for the target language (English) are obtained with \texttt{spaCy}.

\subsection{Models}
\begin{figure}[t]
\centering
\includegraphics[width=14.0cm]{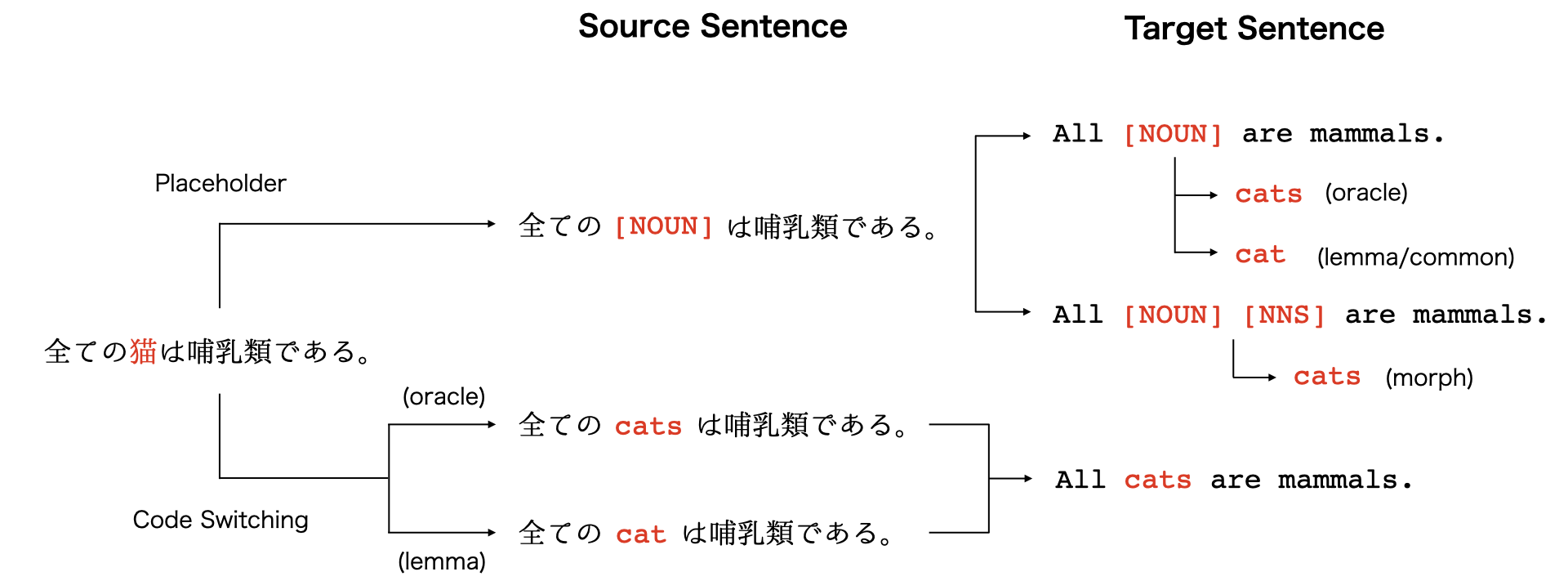}
\caption{The preprocessing of the lexically constrained baseline models.}
\label{fig:baseline}
\end{figure}

As the baseline, we implement a Transformer \citep{NIPS2017_3f5ee243} translation model based on \texttt{AllenNLP} \citep{Gardner2017AllenNLP}.
We configure the model in the Transformer-base setting and sentences are tokenized using \texttt{sentencepiece} \citep{kudo-2018-subword}, which has a shared source-target vocabulary of about 16k sub-words.
The overviews of lexically constrained models are summarized in \fig{fig:baseline}.

\minisection{Placeholder (PH)}
In the placeholder method, the model is trained to translate sentences with a placeholder token and pass that through to the translation. In our experiments, we use different placeholder tokens \texttt{[NOUN]} and \texttt{[VERB]} for nouns and verbs.
Predicted placeholder tokens are replaced by the pre-specified term in the post-processing step.
We evaluate three types of placeholder baselines, each of which differs in what inflected form the target placeholder token is replaced with: {\bf PH (oracle)}, where the pre-specified term is embedded in the same form as in the reference; {\bf PH (lemma)}, always the lemma form; {\bf PH (common)}, the most common inflected forms in the training data, which are the singular form for \texttt{[NOUN]} and the past tense form for \texttt{[VERB]}. The results of PH (lemma) and PH (common) are provided as naive baselines to give a sense of how difficult predicting the correct inflected form is.

We also provide a baseline that performs word inflection through an external resource ({\bf PH (morph)}).
As in \citet{tamchyna-etal-2017-modeling}, words that need inflection are followed by morphological tags, and word formation is realized through an external resource.
We use \texttt{LemmInflect}\footnote{\url{https://github.com/bjascob/LemmInflect}} to decompose the dictionary entries with their lemma and part-of-speech tags and to recover the inflected word form.
As this model uses an external resource to perform inflection, it is not directly comparable with our proposed models but we provide its results as an oracle baseline.

\minisection{Code-switching (CH)}
The code-switching model replaces a phrase in a source sentence with the corresponding target phrase according to a bilingual dictionary.\footnote{\citet{dinu-etal-2019-training} utilize source factors that indicate which tokens are code-switched, but we observe no significant difference by adding source factors. Therefore, we simply report the results from the model with minimal components.}
{\bf CH (oracle)} uses the same target words as in the reference, and {\bf CH (lemma)} uses the lemma form.

\minisection{Proposed Model}
We implement our proposed model described in \cref{sec:approach} on top of the placeholder baseline model.
Compared to the baseline, our proposed model has three additional modules: the target context encoder, target character embeddings, and character-level decoder.
The embedding and hidden sizes are all set to 512, which is the same as in the Transformer-base model.
The additional encoder and decoder have two layers, and the feedforward dimension is 1024.

Note that, for all the models, we restrict the number of constraints to at most one in each sentence as an initial investigation. This favors the placeholder-based models as handling more than one placeholder introduces additional complexity in the system and tends to degrade the performance, while the code-switching methods suffer less from multiple constraints \citep{song-etal-2019-code}.
We leave experiments with multiple constraints to future work.

\subsection{Training with Lexical Constraints}
To apply lexical constraints, the models are trained with data augmentation.
Augmented data is created for all sentences that contain any of the source and target phrases found in the dictionary entries.
To control the amount of augmented data to around 10\% of the original training data, we restrict the dictionary entries to infrequent ones.
The restriction to infrequent phrases also simulates real-word use-cases, where user-specified terms are often rare words that typical NMT models struggle with in translation.
Specifically, we restrict the noun entries to ones with a count at most 20, and the verb entries to 2000.
The threshold is chosen to balance the amount of noun and verb entries in the augmented data.

\subsection{Optimization}
We optimize the models using Adam \citep{Kingma2015AdamAM} with the Noam learning rate scheduler with 8000 warmup steps \citep{NIPS2017_3f5ee243}. The training is stopped when the validation BLEU score does not improve for 3 epochs.

For our proposed model, we found that optimizing the word-level modules and character-level modules separately stabilizes the training process and improves the translation quality.
We first train a normal placeholder model, use the weights to initialize those of our proposed model, and then only update the parameters of the additional modules.
In this second training stage, we use the loss value as validation metric and stop the training when the lowest value is updated for 5 epochs.

\section{Results}
\label{sec:results}
\subsection{Evaluation}
For each model, we evaluate the overall translation quality with BLEU \citep{papineni-etal-2002-bleu}.\footnote{SacreBLEU\citep{post-2018-call} version string: \\\texttt{case.mixed+numrefs.1812+smooth.exp+tok.13a+version.1.5.1}}
We also evaluate the {\it specified term use rate}, a metric to check if the model correctly includes the specified target term.
Note that this is only an approximate measure of what we want to measure: whether the specified term is used in the correct form in the output translation.
Since a single source sentence can be translated into different grammatical constructions, it is possible that the inflected form in the system output is different from the one in the reference but still correct in the context.
Still, we find a substantial overlap in the inflectional form of the specified term between the reference and the system output, and thus report this metric, followed by a more closely inspected manual evaluation.

Also, we are interested in how well the model generalizes to dictionary entries unseen during training. In typical use cases of lexically constrained translation, the specified terms are new or rare words that are not likely to appear in the training data. We construct two kinds of evaluation dictionaries: {\it seen} and {\it unseen}.
We first construct a dictionary by aggregating only entries that appear in the dev/test set.
Then, we randomly split the entries into {\it seen} and {\it unseen} and remove the {\it unseen} entries from the training dictionary. Thus, the {\it seen} split contains entries that appear in the training data while the {\it unseen} not.
We evaluate the model separately using the noun and verb dictionary, which results in a total of four kinds of evaluation configurations.

\begin{table*}[h]
  \centering
  \begin{tabular}{lllll} \toprule
      & \multicolumn{2}{c}{NOUN}                              & \multicolumn{2}{c}{VERB}                              \\
      & \multicolumn{1}{c}{{\it seen}} & \multicolumn{1}{c}{{\it unseen}} & \multicolumn{1}{c}{{\it seen}} & \multicolumn{1}{c}{{\it unseen}} \\ \midrule
      Baseline & 27.1 / 68.3 & 27.1 / 66.5 & 27.1 / 63.4 & 27.1 / 61.2 \\ \midrule
      CS (oracle) & 27.3 / 86.8 & 27.0 / 79.3 & 27.5 / 91.9 & 27.2 / 43.9 \\
      PH (oracle) & 27.2 / 98.8 & 27.0 / 99.2 & 27.4 / 98.7 & 27.5 / 99.4 \\ \midrule
      PH (lemma) & \multirow{2}{*}{27.1 / 84.7}  & \multirow{2}{*}{26.9 / 84.0} & 26.9 / 9.41 & 27.1 / 11.4 \\
      PH (common) &                            &  & 27.3 / 81.8 & 27.3 / 68.9 \\
      CS (lemma) & 27.4 / 81.7 & 27.1 / 74.6 & 27.6 / 81.7 & 27.3 / 42.1 \\
      Proposed & 27.2 / 89.9 & 26.9 / 79.1 & 27.4 / 88.3 & 27.4 / 73.9 \\
      PH (morph) & 27.9 / 84.7 & 27.8 / 81.2 & 28.5 / 91.1 & 28.4 / 87.9 \\
  \bottomrule
  \end{tabular}
\caption{BLUE scores and the specified term use rate of the different models over different evaluation dictionaries. CS: Code-switching, PH: placeholder. For NOUN, PH (lemma) and PH (common) are the same model because the most common inflection for nouns is their lemma.}
\label{table:results-overall}
\end{table*}

\subsection{Main Results}
The results are shown in Table \ref{table:results-overall}. For each configuration, we report the average of three models trained with different random seeds.

First, the lexically constrained models show BLEU scores not significantly different from the baseline.
The only exception is PH (morph): it consistently improves the BLEU score by from 0.7 to 1.4 points from the baseline.
This indicates the strength of injecting the NMT model with morphological knowledge for better generalization in translation.
In the following discussion, we focus on the comparison of the specified term use rate.

PH (oracle) and CS (oracle) models receive the same inflected form of a specified term as in the reference, and thus offer upper bounds for the specified term use rate.
We observe that PH (oracle) exhibits nearly perfect specified term use rates (more than 98\% with all dictionaries).
Also, it is more successful at incorporating the specified term into translation than CS (oracle) in the setting of one constraint, which is in line with previous observations \citep{song-etal-2019-code}.

As for the models that need to handle inflection, the results are quite mixed for NOUN.
A simple strategy of predicting the most common inflection achieves better specified term use rates than most of the other sophisticated models.
We conjecture that some examples allow either singular or plural form and that makes a proper evaluation difficult. Therefore, we turn to the results from VERB for model comparison.

In terms of both {\it seen} and {\it unseen} of the VERB dictionary, PH (morph) performs the best.
Note, however, that this model is not comparable to our model as it assumes access to a high-quality morphological analyzer at training time to obtain morphological tags and the correct inflectional paradigm of user-specified terms at prediction time.

In a more restricted setting, our proposed model outperforms the comparable code-switching model (CS (lemma)) and the other baselines.
In particular, the proposed model is more robust than CS (lemma) to {\it unseen} specified terms: we observe a consistent tendency that the specified term use rate degrades when the entries are unseen during training especially with CS (lemma) and verb entries (81.7 to 42.1), while this tendency is less pronounced in the placeholder model with lemmas (88.3 to 73.9).
Overall, our model exhibits faithfulness to lexical constraints similar to those of the normal placeholder model while having flexibility, which we examine below.

\subsection{Fine-grained Analysis}
The specified term use rate only checks whether specified terms are used in the same form as in the reference. Now we examine the systems' output more closely by manual inspection.
As the problem of inflection matters more in verbs than in nouns in English, here we focus on the translation with the verb dictionary.

We sample from the system's output of the test set 50 sentences with the {\it seen} and {\it unseen} lexical constraints respectively.
We manually check the sampled sentences and annotate each sentence with one of the three tags: {\bf {\it correct}} --- the specified term is used in the translation in the correct inflected form (not necessarily the same as in the reference); {\bf {\it incorrect}} --- the model produces the specified term in some inflected form but that results in an ungrammatical sentence; {\bf {\it null}} --- the model fails to produce the specified term in any form.
The result is shown in Table \ref{table:results-manual}.

\begin{table}[]
  \centering
  \begin{tabular}{lcc} \toprule
                            & VERB {\it seen}  & VERB {\it unseen} \\
  CS (lemma)            & 49 / 0 / 1 & 26 / 0 / 24 \\
  PH with lemmas (proposed) & 48 / 2 / 0 & 39 / 7 / 4  \\
  PH (morph)          & 50 / 0 / 0 & 47 / 3 / 0 \\ \bottomrule
  \end{tabular}
\caption{The manual evaluation of the 50 sampled sentences. The values in each cell indicate {\it correct} / {\it incorrect} / {\it null}.}
\label{table:results-manual}
\end{table}

Firstly, for the words that are seen in the training data, all the models mostly generate the correct word form in the context.
On the other hand, the evaluation with VERB unseen reveals both the advantages and disadvantages of each model, which we discuss with examples below.

\minisection{The placeholder model with morphological tags can handle inflection well}
The model mostly generates the correct inflectional form of the specified terms.
The only three exceptions from VERB seen are errors in choosing the transitive or intransitive usage of the term (Table \ref{fig:ph_pos_wrong}).

\begin{table}[h]
  \centering
  \begin{tabularx}{\textwidth}{X} \toprule
    {\bf Source}: \ja{特発性肺線維症(IPF)患者14例及びIPF急性増悪で\textcolor{red}{入院}した患者8例を対象として,BALF・血漿に関してウィルス検査・免疫血清学的検査を施行した}\\
    {\bf Reference}: The virus inspection and immunoserologic inspection of BALF and blood plasma were carried out for 14 idiopathic pulmonary fibrosis (IPF) patients and of 8 patients \textcolor{red}{hospitalized} for IPF acute aggravation. \\
    {\bf System Output}: Wils inspection and immunoserologic inspection were enforced on BALF blood and blood in 14 patients with idiopathic pulmonary fibrosis (IPF) and 8 patients who \textcolor{red}{hospitalized} in the IPF acute aggravation. \\ \bottomrule
  \end{tabularx}
\caption{A translation example with the placeholder model with morphological tags. The system output should have generated {\it were hospitalized} in the red part.}
\label{fig:ph_pos_wrong}
\end{table}

\minisection{The code-switching method always produces grammatical inflectional forms}
We observe no {\it incorrect} examples from the code-switching model.
Since the output is determined solely by the word decoder with no additional post-editing performed, if the word decoder is well trained, we can expect the output sentences to be grammatical.

\minisection{The code-switching method tends to fail to observe the constraints}
However, the code-switching methods fail to produce the specified term in 24 examples out of 50, which is notably higher than the other methods.
A typical error is the model ignoring the constraint and producing a synonym, for example, generating {\it conclude} instead of {\it judge}, {\it examine} instead of {\it study}.
This is reasonable given the model architecture.
A well-trained NMT model usually assigns similar vector representations to synonyms.
Even when the specified term is given in the source sentence, it is given a representation similar to other synonyms inside the model, and thus the decoder can generate any words with similar meaning.
We also observe a few character decoding errors: wrongly generating {\it hot-spitalized} instead of {\it hospitalized}, {\it move} instead of {\it remove}.

\minisection{The placeholder method almost always produces the specified term, but sometimes fails to inflect it correctly}
The placeholder method fails to observe the constraint much less frequently than the code-switching method (only 4 examples out of 50).
In most cases (39 examples out of 50), the model can successfully predict the correct form as shown in Table \ref{fig:ph_example_correct}.

\begin{table}[h]
  \centering
  \begin{tabularx}{\textwidth}{X} \toprule
    {\bf Source}: \ja{フローセンサーの原理は浮遊式流量計のテーパー管内フロートの位置を差動トランスで検出し,これの電圧制御により流量を\textcolor{red}{管理}する。}\\
    {\bf Reference}: The sensor controls the flow rate by detecting the position of the float in the tepered tube with a differential transformer and \textcolor{red}{controlling} it with the obtained voltage. \\
    {\bf System Output}: The principle of the flow sensor is that the position of the float in the taper tube of the floating flowmeter is detected by the differential transformer, and the flow rate is \textcolor{red}{controlled} by this voltage control. \\ \bottomrule
  \end{tabularx}
  \caption{A translation example with the placeholder model with a character decoder. The model predicts the correct inflectional form of {\it control} that fits in the context.}
  \label{fig:ph_example_correct}
\end{table}

The failures consist of generalization errors of inflectional form: generating {\it maken} for {\it make}.
It is impossible in principle to correctly predict irregular inflectional forms that are unseen in the training data, but this is usually not much of a problem since the specified term is usually a rare or new word, which tends to have a regular inflectional paradigm.
The other kind of error we observe is the model predicting a well-defined word form that is wrong in the context (Table \ref{fig:ph_example_wrong}). We expect that both error types can be addressed by exploiting additional data, either parallel or monolingual, to learn inflection rules in the target language.

\begin{table}[t]
  \centering
  \begin{tabularx}{\textwidth}{X} \toprule
    {\bf Source}: \ja{国立病院機構関門医療センター(国立下関病院)は2002年9月30日に女性総合診療を\textcolor{red}{開設}した。}\\
    {\bf Reference}: A National Hospital System Kanmon Medical Center (A National Shimonoseki Hospital) \textcolor{red}{opened} the comprehensive woman medical care service on September 30th in 2002. \\
    {\bf System Output}: National Hospital Mechanism Kanmon Medical Center ( the national Shimonoseki Hospital ) \textcolor{red}{opening} the woman general medical care on September 30th, 2002. \\ \bottomrule
  \end{tabularx}
\caption{A translation example with the placeholder model with a character decoder. The model predicts a wrong inflectional form for {\it open}.}
\label{fig:ph_example_wrong}
\end{table}

\section{Conclusion and Future Work}
In this study, we point out that the traditional placeholder translation method embeds the specified term into the generated translation without considering the context of the placeholder token, which potentially leads to grammatically incorrect translations.
To address this shortcoming, we proposed a flexible placeholder translation model that handles inflection when the specified term is given in the form of a lemma.
In the experiment of the Japanese-to-English translation task, we showed that the proposed model can inflect user-specified terms more accurately than the code-switching method.

Future work includes testing the proposed method on morphologically-rich languages or extending the model to handle more than one placeholder in a sentence.
Also, the proposed model still has room for improvement to learn inflection.
It is possible that we can improve the model by exploiting monolingual corpora in the target language to provide additional training signals for learning the correct inflection in context.

\section*{Acknowledgements}
The authors would like to thank the anonymous reviewers for their helpful comments.
This work was supported by the Research and Development of Deep Learning Technology for Advanced Multilingual Speech Translation, the Commissioned Research of National Institute of Information and Communications Technology (NICT), JAPAN.

\small

\bibliographystyle{apalike}
\bibliography{sections/reference}

\end{document}